\documentclass{acm_proc_article-sp}

\usepackage{fixltx2e}
\usepackage{amssymb}
\usepackage{amsmath}
\usepackage{amsfonts}
\usepackage{hyphenat}

\usepackage[usenames,dvipsnames]{xcolor}
\usepackage[normalem]{ulem}	

\usepackage{hyperref}
\hypersetup{
    bookmarks=true,		
    bookmarksopen=true,	
    unicode=true,	   	
    pdftitle={Applying Strategic Multiagent Planning to Real-World Travel Sharing Problems}, 
    pdfauthor={Jan Hrncir, Michael Rovatsos},	
    pdfsubject={ATT Workshop Paper},				
    pdfkeywords={multiagent planning, real-world application, travel sharing}, 
    colorlinks=true, 	
    linkcolor=darkblue,	
    citecolor=darkblue,	
    filecolor=magenta,	
    urlcolor=cyan		
}

\usepackage{color}
\usepackage{framed}

\begin{document}

\newcommand{\mr}[1]{\begin{color}{red}#1\end{color}} 
\newcommand{\mrc}[1]{\begin{color}{OliveGreen}\footnote{#1}\end{color}}
\newcommand{\mrd}[1]{\begin{color}{red}\sout{#1}\end{color}}

\newcommand{\jh}[1]{#1} 
\newcommand{\jhh}[1]{#1} 
\newcommand{\jhhh}[1]{#1} 
\newcommand{\jhc}[1]{} 
\newcommand{\jhd}[1]{} 
\newcommand{\todo}[1]{{\bf {\color{Maroon} TODO:} #1}}

\definecolor{shadecolor}{rgb}{0.93,0.93,0.93} 
\def\settingsForEnum{
    \setlength{\topsep}{0pt}
    \setlength{\partopsep}{0pt}
    \setlength{\itemsep}{0pt}
    \setlength{\parskip}{0pt}
    \setlength{\parsep}{0pt}
}

\def\sgp          {SGPlan\textsubscript{6}}
\def\sgpfive      {SGPlan\textsubscript{5}}
\def\Astar        {A\textsuperscript{*}}
\def\sas          {SAS\textsuperscript{+}}

\definecolor{shadecolor}{rgb}{0.90,0.90,0.90}
\definecolor{darkblue}  {rgb}{0,0,0.7}
\definecolor{darkgreen} {rgb}{0,0.7,0}
\definecolor{cyan}      {rgb}{0,0.5,0.6}

\title{Applying Strategic Multiagent Planning to Real-World Travel Sharing Problems}



%
%
%
%

%

\numberofauthors{2}

\author{
%
\alignauthor
Jan Hrn\v{c}\'{i}\v{r}\\
       \affaddr{Agent Technology Center}\\
       \affaddr{Faculty of Electrical Engineering}\\
       \affaddr{Czech Technical University in Prague}\\
       \affaddr{121 35 Prague, Czech Republic}\\
       \email{jan.hrncir@agents.fel.cvut.cz}
\alignauthor
Michael Rovatsos\\
       \affaddr{School of Informatics}\\
       \affaddr{The University of Edinburgh}\\
       \affaddr{Edinburgh EH8 9AB, United Kingdom}\\
       \email{mrovatso@inf.ed.ac.uk}
}

\maketitle

\begin{abstract}
Travel sharing, i.e., the problem of finding parts of routes which can be
shared by several travellers with different points of departure and destinations, is
a~complex multiagent problem that requires taking into account individual
agents' preferences to come up with mutually acceptable joint plans. In this paper,
we apply state-of-the-art planning techniques to real-world public transportation
data to evaluate the feasibility of multiagent planning techniques in this
domain. The potential of improving travel sharing technology
has great application value due to its ability to reduce the 
environmental impact of travelling while providing benefits to travellers at
the same time.

We propose a three-phase algorithm that utilises performant single-agent
planners to find individual plans in a simplified domain first and then merges them
using a best-response planner which ensures resulting solutions are individually
rational. Finally, it maps the resulting plan onto the full temporal planning
domain to schedule actual journeys.


The evaluation of our algorithm on real-world, multi-mod\-al public
transportation data for the United Kingdom shows linear scalability both in the
scenario size and in the number of agents, where trade-offs have to be made
between total cost improvement, the percentage of feasible timetables identified
for journeys, and the prolongation of these journeys. Our system constitutes the
first implementation of strategic multiagent planning algorithms in large-scale
domains and provides insights into the engineering process of translating
general domain-independent multiagent planning algorithms to real-world
applications.




\end{abstract}










\section{Introduction}

Travelling is an important and frequent activity, yet people willing to travel
have to face problems with rising fuel prices, carbon footprint and traffic
jams. These problems can be ameliorated by {\em travel sharing}, i.e., groups of
people travel together in one vehicle for parts of the journey. Participants in
such schemes can benefit from travel sharing in several ways: sharing parts of
a journey may reduce cost (e.g., through group tickets), carbon footprint (e.g.,
when sharing a private car, or through better capacity utilisation of public
means of transport), and travellers can enjoy the company of others on a long
journey. In more advanced scenarios one could even imagine this being combined
with working together while travelling, holding meetings on the road, etc.

Today, there exist various commercial online services for {car}\footnote{E.g.,
\href{https://www.liftshare.com/uk/}{www.liftshare.com} or
\href{http://www.citycarclub.co.uk/}{www.citycarclub.co.uk}.}, bike, and walk
sharing as well as services which assist users in negotiating shared
journeys\footnote{E.g.,
\href{http://www.companions2travel.co.uk/}{www.companions2travel.co.uk},
\href{http://www.travbuddy.com/}{www.travbuddy.com}.}, and, of course, plenty of
travel planning services\footnote{E.g., in the United Kingdom:
\href{http://www.nationalrail.co.uk/}{www.nationalrail.co.uk}~for
trains,~\href{http://www.traveline.info/}{www.traveline.info}
and~\href{http://www.google.com/transit}{www.google.com/transit} for~multi-modal
transportation.} that automate {\em individual} travel planning for one or
several means of transport. On the research side, there is previous work
that deals with the ridesharing and car-pooling problem \cite{abdel2007,
ferrari2003, lalos2009}.
However, no work has been done that attempts to compute {\em joint} travel
plans based on {\em public transportation} time\-tab\-le data and geographical
stop locations, let alone in a~way that takes into account the {\em strategic}
nature of the problem, which comes about through the different (and potentially
conflicting) preferences of individuals who might be able to benefit from
travelling together. From the point of view of (multiagent) planning, this
presents itself as a very complex application scenario: Firstly, even if one
restricted oneself to centralised (non-strategic) planning, the domain is huge
-- public transportation data for the UK alone currently involves $240,590$
timetable connections for trains and coaches (even excluding local city buses),
which would have to be translated to a quarter of a million planning actions, at
least in a naive formalisation of the domain. Secondly, planning for multiple
self-interested agents that are willing to cooperate only if it is beneficial
for them is known to be exponentially harder than planning for each agent
individually~\cite{brafman2008}. Yet any automated service that proposes joint
journeys would have to guarantee such {\em strategic} properties in order to be
acceptable for human users (who could then even leave it to the service to
negotiate trips on their behalf).


In this paper, we present an implementation of best-res\-pon\-se planning (BRP)
\cite{jonsson2011} within a three-phase algorithm that is capable of solving
strategic travel sharing problems for several agents based on real-world
transportation data. Based on a simplified version of the domain that ignores
timetabling information, the algorithm first builds individual travel plans
using state-of-the-art single-agent planners that are available off the shelf.
It then merges these individual plans and computes a multiagent plan that is a
Nash equilibrium and guarantees individual rationality of solutions, as well as
stability in the sense that no single agent has an incentive to deviate from the
joint travel route. This is done using BRP as the underlying planner, as it is
the only available planner that can solve strategic multiagent planning problems
of such scale, and is proven to converge in domains that comply with certain
assumptions, as is the case for our travel sharing domain. In a third and final
step, the resulting multiagent plan is mapped onto the full temporal planning
domain to schedule actual journeys. This scheduling task is not guaranteed to
always find a feasible solution, as the previous simplification ignores a
potential lack of suitable connections. However, we show through an extensive
empirical evaluation that our method finds useful solutions in a large number of
cases despite its theoretical incompleteness.

The contribution of our work is threefold: Firstly, we show that current
multiagent planning technology can be used in important planning domains such as
travel sharing by presenting its application to a practical problem that cannot
be solved with other existing techniques. In the process, we describe the
engineering steps that are necessary to deal with the challenges of real-world
large-scale data and propose suitable solutions. Secondly, we present an
algorithm that combines different techniques in a practically-oriented way, and
which is largely based on domain-independent off-the-shelf heuristic problem
solvers. In fact, only data preprocessing and timetable mapping use
domain-specific knowledge, and much of the process of incorporating this
knowledge could be replicated for similar other domains (such as logistics,
manufacturing, and network communications).
Finally, we provide a potential solution to the hard computational problem of
travel sharing that could be exploited for automating important tasks in a
future real-world application to the benefits of users, who normally have to
plan such routes manually and would be overwhelmed by the choices in a domain
full of different transportation options which is inhabited by many potential
co-travellers.

We start off by describing the problem domain in section~\ref{sec:domain} and
specifying the planning problem formally in section~\ref{sec:problem}, following
the model used in~\cite{jonsson2011}. Section~\ref{sec:algorithm} introduces our
three-phase algorithm for strategic planning in travel sharing domains and we
present an extensive experimental evaluation of the algorithm in
section~\ref{sec:evaluation}. Section~\ref{sec:discussion} presents
a~discussion of our results and section \ref{sec:conclusion} concludes.

\section{The travel sharing domain} \label{sec:domain}

The real-world travel domain used in this paper is based on the public transport
network in the United Kingdom, a~very large and complex domain which contains
$4,055$ railway and coach stops supplemented by timetable information. An agent
representing a~passenger is able to use different means of transport during its
journey: walking, trains, and coaches. The aim of each agent is to get from its
starting location to its final destination at the lowest possible cost, where
the cost of the journey is based on the duration and the price of the journey.
Since we assume that all agents are travelling on the same day and that
all journeys must be completed within 24~hours, in what follows below we
consider only travel data for Tuesdays (this is an arbitrary choice that could
be changed without any problem).
For the purposes of this paper, we will make the assumption that sharing a~part
of a~journey with other agents is cheaper than travelling alone. While this may
not currently hold in many public transportation systems, defining hypothetical
cost functions that reflect this would help assess the potential benefit of
introducing such pricing schemes.


\subsection{Source data}
\label{sec:sourceData}


The travel sharing domain uses {the National Public Transport Data Repository
(NPTDR)}\footnote{\href{http://data.gov.uk/dataset/nptdr}{data.gov.uk/dataset/nptdr}}
which is publicly available from the Department for Transport of the British
Government. \jh{It contains a~snapshot of route and timetable data} that has
been gathered in the first or second complete week of October since 2004. For
the evaluation of the algorithm in section~\ref{sec:evaluation}, we used data
from
2010\footnote{\href{http://www.nptdr.org.uk/snapshot/2010/nptdr2010txc.zip}{www.nptdr.org.uk/snapshot/2010/nptdr2010txc.zip}},
which is provided in {TransXChange XML}\footnote{An XML-based UK standard for
interchange of route and timetable data.}.


{National Public Transport Access Nodes
(NaPTAN)}\footnote{\href{http://data.gov.uk/dataset/naptan}{data.gov.uk/dataset/naptan}}
is a~UK national system for uniquely identifying all the points of access to
public transport. \jh{Every point of access (bus stop, rail station, etc.) is
identified by an ATCO code\footnote{A~unique identifier for all points of access
to public transport in the United Kingdom.}, e.g., {\em 9100HAYMRKT} for
Haymarket Rail Station in Edinburgh.} Each stop in NaPTAN XML data is also
supplemented by common name, latitude, longitude, address and other pieces of
information. This data also contains information about how the stops are grouped
together (e.g., several bus bays that are located at the same bus station).

\begin{figure}
\centering
\includegraphics[width=77mm]{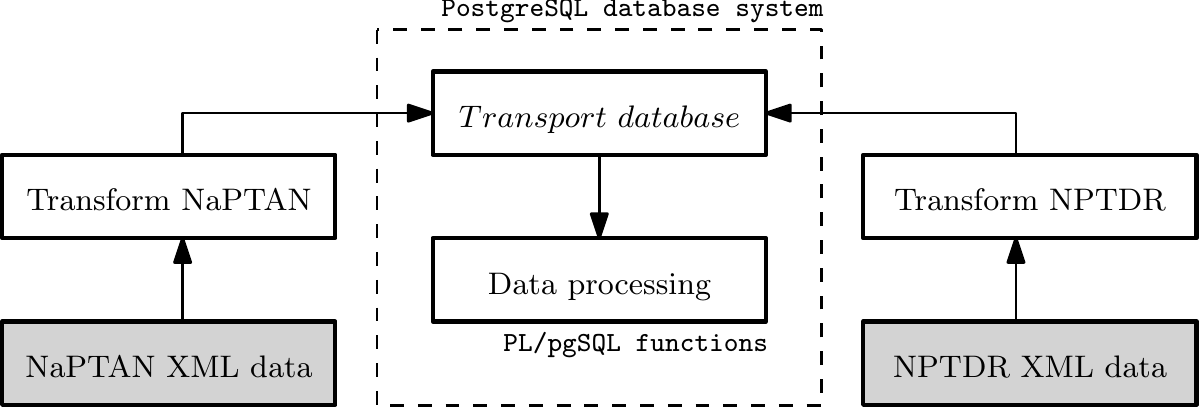}
\caption{Overview of the data transformation and processing}
\label{fig:dataDiagram}
\end{figure} 

To be able to use this domain data with modern AI planning systems, it has to
be converted to the Planning Domain Definition Language (PDDL). We transformed the
data in three subsequent stages, cf.~Figure~\ref{fig:dataDiagram}. First, we
transformed the NPTDR and NaPTAN XML data to a~spatially-enabled PostgreSQL
database. Second, we automatically processed and optimised the data in the
database. The data processing by SQL functions in the procedural PL/pgSQL
language included the following steps: merging bus bays at bus stations and parts of
train stations, introducing walking connections to enable multi-modal journeys, 
and eliminating duplicates from the timetable. Finally, we created a~script for
generating PDDL specifications based on the data in the database. More details
about the data processing and PDDL specifications can be found in
\cite{hrncir2011}.


\begin{figure}
\centering
\includegraphics[width=70mm]{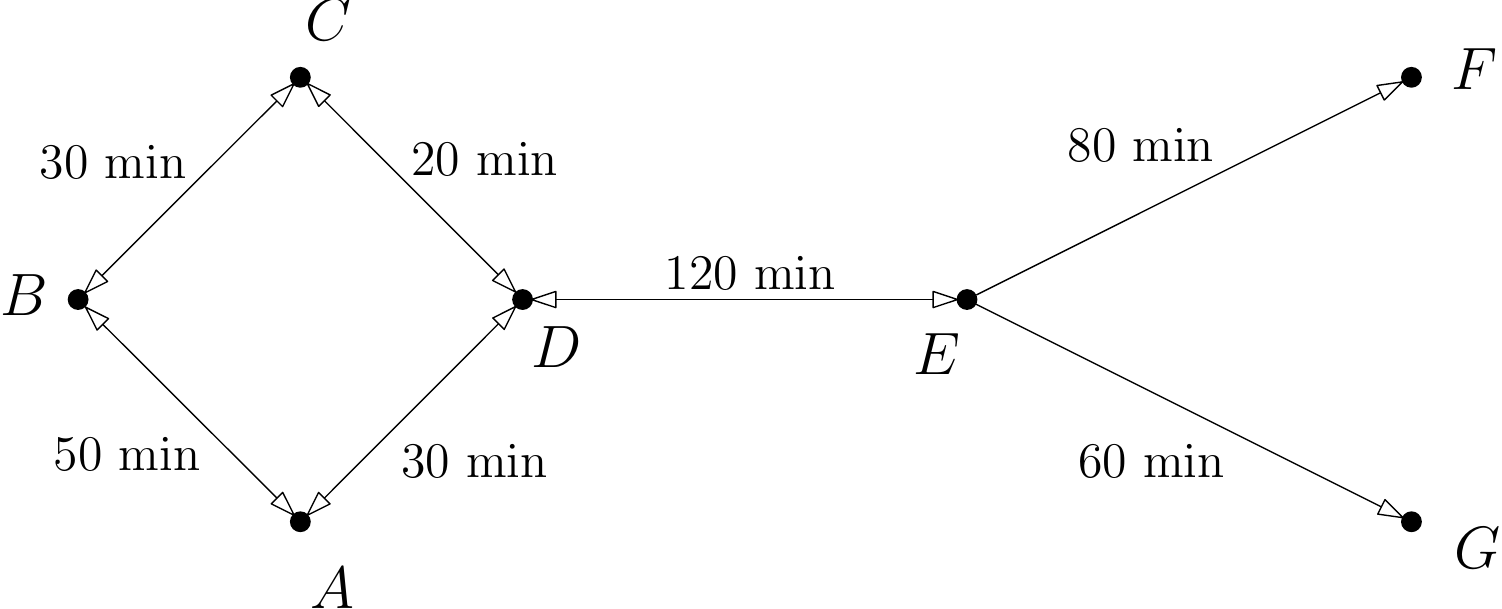}
\caption{An example of the relaxed domain (e.g., it takes 50~minutes to travel
from the stop $\boldsymbol{A}$~to~$\boldsymbol{B}$) }
\label{fig:relaxedWithTime}
\end{figure}

\subsection{Planning domain definitions}
\label{sec:domainDefinitions}

Since the full travel planning domain is too large for any current
state-of-the-art planner to deal with, we distinguish the {\em full domain} from
a {\it relaxed domain}, which we will use to come up with an initial plan before
mapping it to the full timetable information in our algorithm below.

The {\it relaxed domain} is a~single-agent planning domain represented as
a~directed graph where the nodes are the stops and the edges are the connections
provided by a~service. The graph must be directed because there exist stops that
are used in one direction only. There is an edge from $A$~to~$B$ if there is at
least one connection from $A$~to~$B$ in the timetable. The cost of this edge
is the minimal time needed for travelling from $A$~to~$B$. A~plan~$P_i$ found in
the relaxed domain for the agent $i$ is a~sequence of connections to travel from
its origin to its destination. \jhhh{The relaxed domain does not contain any
information about the traveller's departure time. This could be problematic in a scenario where people are
travelling at different times of day. This issue could be solved by clustering
of user requests, cf.~chapter~\ref{sec:conclusion}.}



A~small example of the relaxed domain is shown in
Figure~\ref{fig:relaxedWithTime}. An example plan for an agent travelling from
$C$~to~$F$ is $P_1 = \langle C \rightarrow D, D \rightarrow E, E \rightarrow F
\rangle$. To illustrate the difference between the relaxed domain and the full
timetable, \jh{there are $8,688$ connections in the relaxed domain for trains
and coaches in the UK compared to $240,590$ timetable connections.}

Direct trains that do not stop at every stop are filtered out from the
relaxed domain for the following reason: Assume that in
Figure~\ref{fig:relaxedWithTime}, there is only one agent travelling from
$C$~to~$F$ and that its plan in the relaxed domain is to use a~direct train from
$C$~to~$F$. In this case, it is only possible to match its
plan to direct train connections from $C$~to~$F$, and not to trains that stop at
$C$,~$D$,~$E$, and $F$. Therefore, the agent's plan cannot be matched against
all possible trains between $C$ and $F$ which is problematic especially in the
case where the majority of trains stop at every stop and only a~few trains are
direct. On the other hand, it is possible to
match a~plan with a~train stopping in every stop to a~direct train, as it is
explained later in section~\ref{sec:timetabling}.

The {\it full domain} is a~multiagent planning domain based on
the joint plan~$P$. Assume that there are $N$~agents in the full domain (each
agent~$i$ has the plan~$P_i$ from the relaxed domain). Then, the joint plan~$P$
is a~merge of single-agent plans defined by formula 
\begin{equation} \label{eq:defJointPlan} P = \bigcup_{i=1}^{N} P_i
\nonumber \end{equation} where we interpret $\bigcup$ as the union of graphs
that would result from interpreting each plan as a set of edges connecting stops. 
More specifically, given a~set of single-agent plans, the plan merging operator~$\bigcup$
computes its result in three steps: First, it transforms every single-agent
plan $P_i$ to a~directed graph~$G_i$ where the nodes are the stops from the
single-agent plan $P_i$ and the edges represent the actions of $P_i$ (for
instance, a~plan $P_1 = \langle C \rightarrow D, D \rightarrow E, E \rightarrow
F \rangle$ is transformed to a~directed graph $G_1 = \{ C \rightarrow D
\rightarrow E \rightarrow F \}$). Second, it performs a~graph union operation
over the directed graphs $G_i$ and labels every edge in the joint plan
with the numbers of agents that are using the edge (we don't introduce any
formal notation for these labels here, and simply slightly abuse the standard
notation of sets of edges to describe the resulting graph).

\begin{figure}
\centering
\includegraphics[width=70mm]{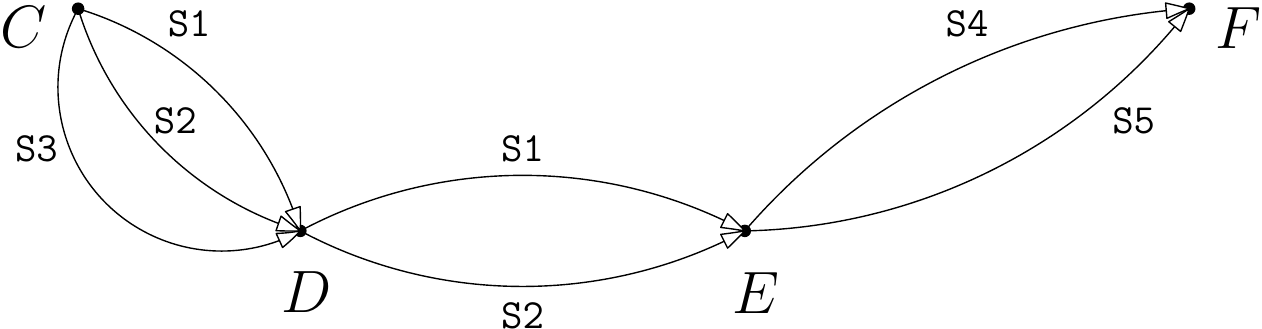}
\caption{An example of the full domain with stops $\boldsymbol{C}$,
$\boldsymbol{D}$, $\boldsymbol{E}$ and $\boldsymbol{F}$ 
for the joint plan~$\boldsymbol{P} \boldsymbol{=} \boldsymbol\{
\boldsymbol{C} \xrightarrow{(1)}
\boldsymbol{D} \xrightarrow{(1, 2)} \boldsymbol{E} \xrightarrow{(1)}
\boldsymbol{F} \boldsymbol\}$}
\label{fig:fullDomain}
\end{figure}

As an example, the joint plan for agent 1 travelling from $C$~to~$F$ and sharing
a journey from $D$ to $E$ with agent 2
would be computed as
\begin{multline*} \label{eq:exampleJointPlan}
\langle C \rightarrow D, D \rightarrow E,
E \rightarrow F \rangle \: \cup \: \langle D \rightarrow E \rangle =  \\ 
 \{ C \xrightarrow{(1)} D \xrightarrow{(1, 2)} E \xrightarrow{(1)} F \}
\end{multline*}
With this, the {\it full domain} is represented as a~directed multigraph where the
nodes are the stops that are present in the joint plan of the relaxed domain. Edges of the
multigraph are the service journeys from the timetable. Every service is
identified by a~unique service name and is assigned a~departure time from each
stop and the duration of its journey between two stops. In the example of the full
domain in Figure~\ref{fig:fullDomain}, the agents can travel using some subset of five different
services \verb|S1| to \verb|S5|. In order to travel from $C$ to $D$ using 
service \verb|S1|, an~agent must be present at stop $C$ before its
departure.

\section{The planning problem} \label{sec:problem}
Automated planning technology \cite{ghallab2004} has developed a variety of
scalable heuristic algorithms for tackling hard planning problems, where plans,
i.e., sequences of actions that achieve a given goal from a given initial state,
are calculated by domain-independent problem solvers.
To model the travel sharing problem, we use a multiagent planning formalism
which is based on MA-STRIPS~\cite{brafman2008} and coalition-planning
games~\cite{brafman2009}. States are represented by sets of ground fluents,
actions are tuples $a = \langle \mathit{pre}(a), \mathit{eff}(a) \rangle $.
After the execution of action $a$, positive fluents $p$ from $\mathit{eff}(a)$
are added to the state and negative fluents $\neg p$ are deleted from the state.
Each agent has individual goals and actions with associated costs. There is no
extra reward for achieving the goal, the total utility received by an agent is
simply the inverse of the cost incurred by the plan executed to achieve the
goal.

More formally, following the notation of \cite{jonsson2011}, a multiagent
planning problem (MAP) is a tuple $$\Pi=\langle
N,F,I,\{G_i\}_{i=1}^n,\{A_i\}_{i=1}^n,\Psi, \{c_i\}_{i=1}^{n}\rangle$$ where
\begin{itemize}
\item $N = \{1,\ldots,n\}$ is the set of agents,
\item $F$ is the set of fluents, 
\item $I\subseteq F$ is the initial state,
\item $G_i\subseteq F$ is agent $i$'s goal, 
\item $A_i$ is agent $i$'s action set,
\item $\Psi:A\rightarrow\{0,1\}$ is an admissibility function,
\item $c_i:\times_{i=1}^n A_i\rightarrow\mathbb{R}$ is  the cost
function of agent $i$. 
\end{itemize}
$A=A_1\times\ldots\times A_n$ is the joint action set
assuming a concurrent, synchronous execution model, and $G=\wedge_i
G_i$ is the conjunction of all agents' individual goals. A MAP typically
imposes concurrency constraints regarding actions that 
cannot or have to be performed concurrently by different agents to
succeed which the authors of~\cite{jonsson2011} encode using an
admissibility function $\Psi$, with $\Psi(a)=1$ if the joint action
$a$ is executable, and $\Psi(a)=0$ otherwise.


A {\em plan} $\pi=\langle a^1,\ldots,a^k\rangle$ is a sequence of
joint actions $a^j\in A$ such that $a^1$ is applicable in the initial
state $I$ (i.e., $\mathit{pre}(a^1)\subseteq I$), and $a^j$ is applicable
following the application of $a^1, \ldots,$ $a^{j-1}$.
We say that $\pi$ {\em solves} the MAP $\Pi$ if the goal state $G$ is
satisfied following the application of all actions in $\pi$ in sequence.
The cost of a plan $\pi$ to agent $i$ is
given by $C_i(\pi)=\sum_{j=1}^k c_i(a^j)$.
Each agent's contribution to a plan $\pi$ is denoted by $\pi_i$ (a~sequence of
$a_i\in A_i$).
\jhd{, and $\pi_{-i}$ is the joint plan of all remaining agents.}

\subsection{Best-response planning}

The {\em best-response planning} (BRP) algorithm proposed in \cite{jonsson2011}
is an algorithm which, given a~solution $\pi^k$ to a MAP $\Pi$,
finds a solution $\pi^{k+1}$ to a {\em transformed planning problem}
$\Pi_i$ with minimum cost $C_i(\pi^{k+1})$ among all possible solutions:
\ $$\pi^{k+1} = \arg\min \{C_i(\pi)|\pi \textrm{ identical to } \pi^k
\textrm{ for all } j\neq i\}$$ \jh{The transformed planning problem $\Pi_i$ is
obtained} by rewriting the original problem $\Pi$ so that all other agents'
actions are fixed, and agent $i$ can only choose its own actions in such a way that all
other agents still can perform their original actions. Since $\Pi_i$ is a
single-agent planning problem, any cost-optimal planner can be used as a
best-response planner.

In~\cite{jonsson2011}, the authors show \jhhh{how for a class of congestion
planning problems}, where all fluents are {\em private}, the transformation they
propose allows the algorithm to converge to a Nash equilibrium if agents iteratively
perform best-response steps using an optimal planner. This requires that every
agent can perform its actions without requiring another agent, and hence can
achieve its goal in principle on its own, and conversely, that no agent can
invalidate other agents' plans. Assuming infinite capacity of vehicles,
the relaxed domain is an instance of a congestion planning {problem}\footnote{
Following the definition of a congestion planning problem in \cite{jonsson2011},
all actions are private, as every agent can use transportation means on their
own and the other agents' concurrently taken actions only affect action cost.
A part of the cost function defined in~section~\ref{sec:costFunctions} depends
only on the action choice of individual agent.}.




The BRP planner works in two phases: In the first phase, an initial plan for
each agent is computed \jh{(e.g., each agent plans independently or
a~centralised multi-agent planner is used)}. In the second phase, the planner
solves simpler best-response planning problems from the point of view of each
individual agent. The goal of the planner in a~BRP problem is to minimise the
cost of an agent's plan without changing the plans of others. Consequently, it
optimises a~plan of each agent with respect to the current joint plan.

This approach has several advantages. It supports full concurrency of actions
and the BRP phase avoids the exponential blowup in the action space resulting in
much improved scalability. For the class of potential games \cite{monderer1996},
it guarantees to converge to a~Nash equilibrium.  On the other hand, it does not
guarantee the optimality of a~solution, i.e., the quality of the equilibrium in
terms of overall efficiency is not guaranteed (it depends on which initial plan
the agents start off with). However, experiments have proven that it can be
successfully used for improving general multiagent plans \cite{jonsson2011}.
Such non-strategic plans can be computed using a~{\it centralised multiagent
planner}, i.e., a~single-agent planner (for instance Metric-FF
\cite{hoffmann2001}) which tries to optimise the value of the joint cost function (in our case the
sum of the values of the cost functions of agents in the environment) while
trying to achieve all agents' goals. Centralised multiagent planners have no
notion of self-interested agents, i.e., they ignore the individual preferences
of agents.

\section{A three-phase strategic travel sharing algorithm} \label{sec:algorithm}
The main problem when planning for multiple agents with a~centralised
multiagent planner is the exponential blowup in the action space which is
caused by using concurrent, independent actions \cite{jonsson2011}. 
Using a naive PDDL translation has prov\-en that a~direct
application of a~centralised multiagent planner to this problem does not scale
well. For example, a~simple scenario with two agents, ferries to Orkney Islands
and trains in the area between Edinburgh and Aberdeen resulted in a~one-day
computation time.

As mentioned above, we tackle the complexity of the domain by breaking down the
planning process into different phases that avoid dealing with the full
fine-grained timetable data from the outset.
Our algorithm, which is shown in Figure~\ref{fig:algorithmPseudocode}, is
designed to work in three phases. 

\subsection{The initial phase}

First, in the {\it initial phase}, an initial journey is found for each agent
using the relaxed domain. A~journey for each agent is calculated independently
of other agents in the scenario using a~single-agent planner. As a~result, each
agent is assigned a~single-agent plan which will be further optimised in the
next phase. This approach makes sense in our domain because the agents do not
need each other to achieve their goals and they cannot invalidate each other's
plans. 

\begin{figure}[t]
	\begin{shaded}
    	{\bf Input}
    	\begin{itemize}
      		\settingsForEnum
			\item a~relaxed domain
			\item a~set of $N$ agents $A = \{a_1, \dots, a_N\}$
			\item an origin and a destination for each agent
		\end{itemize}

	    {\bf 1. The initial phase}
	    \begin{description}
	        \settingsForEnum
	        \item \quad{\bf For} $i = 1, \dots , N$ {\bf do}
	        \begin{description}
	          \settingsForEnum
	          \item \quad$\,$ Find an initial journey for agent $a_i$ using \\
	          \hbox{a~single-agent} planner.
	        \end{description}
	    \end{description}
	
		{\bf 2. The BR phase} 
	    \begin{description}
	       \settingsForEnum
	       \item \quad{\bf Do until} no change in the cost of the joint plan
	       \smallskip
	       \item \quad\quad{\bf For} $i = 1, \dots , N$ {\bf do} 
	       \begin{enumerate}
	          \settingsForEnum
	          \item Create a simpler best-response planning (BRP) \\
	          		problem from~the~point of view of agent $a_i$.
	          \item Minimise the cost of $a_i$'s plan without changing \\
	          		the plans of others.
	       \end{enumerate}
	       \item \quad\quad{\bf End} 
	    \end{description}
	    
	    {\bf 3. The timetabling phase}
	    \begin{description}
	       \settingsForEnum
	       \item \quad Identify independent groups of agents $G = \{g_1, \dots,
	       g_M\}$. \smallskip
	       \item \quad{\bf For} $i = 1, \dots , M$ {\bf do} 
	       \begin{enumerate}
	          \settingsForEnum
	          \item Find the relevant timetable for group $g_i$.
	          \item Match the joint plan of $g_i$ to timetable using a~temporal
	          		single-agent planner in the full domain with the relevant
	          		timetable.
	       \end{enumerate}
	       \item \quad{\bf End} 
	    \end{description}
  	\end{shaded}
	\caption{Three-phase algorithm for finding shared journeys for agents}
	\label{fig:algorithmPseudocode}
\end{figure}

\subsection{The BR phase}

Second, in the {\it BR phase} (best-response phase), which is also based on the
relaxed domain, the algorithm uses the BRP algorithm as described above.
It~iteratively creates and solves simpler best-response planning problems from
the point of view of each individual agent. In the case of the relaxed domain,
the BRP problem looks almost the same as a~problem of finding a~single-agent
initial journey. The difference is that the cost of travelling is smaller when
an~agent uses a~connection which is used by one or more other agents, as will be
explained below, cf.~equation~\eqref{eq:defGroupCost}.

Iterations over agents continue until there is no change in the cost of the
joint plan between two successive iterations. This means that the joint plan
cannot be further improved using the best-response approach. The output of the
BR phase is the joint plan $P$ in the relaxed domain (defined in
section~\ref{sec:domainDefinitions}) that specifies which connections the agents
use for their journeys and which segments of their journeys are shared. The
joint plan $P$ will be matched to the timetable in the final phase of the
algorithm.



\subsection{The timetabling phase} \label{sec:timetabling}
In the final {\it timetabling phase}, the optimised shared journeys are
matched against timetables using a~temporal single-agent planner which assumes
the full domain.
\jh{For this, as a~first step, independent groups of agents with respect to
journey sharing are identified. An independent group of agents is defined as an edge
disjoint subgraph of the joint plan~$P$. This means that actions of
independent groups do not affect each other so it is possible to find
a~timetable for each independent group separately.}

\begin{figure}
\centering
\includegraphics[width=77mm]{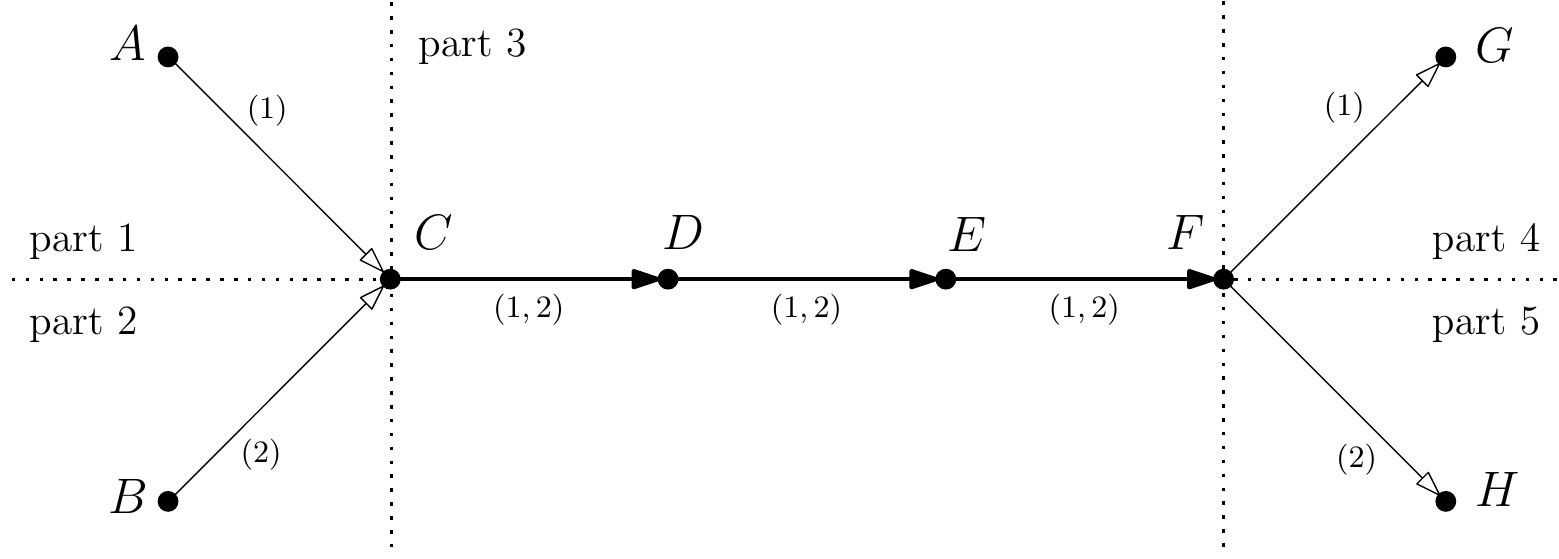}
\caption{Parts of the group journey of two agents}
\label{fig:groupPlanParts}
\end{figure}

Then, for every independent group, {\em parts} of the group
journey are identified. A~{\it part} of the group journey is defined as a~maximal
continuous segment of the group journey which is performed by the same set of agents.
As an example, there is a~group of two agents that share a~segment of their
journeys in Figure~\ref{fig:groupPlanParts}: Agent~1 travels from
 $A$ to $G$ while agent~2 travels from $B$ to $H$. Their group journey
has five parts, with the shared part (part 3) of their journey occurring between stops $C$
and~$F$.

\begin{figure}
\centering
\includegraphics[width=77mm]{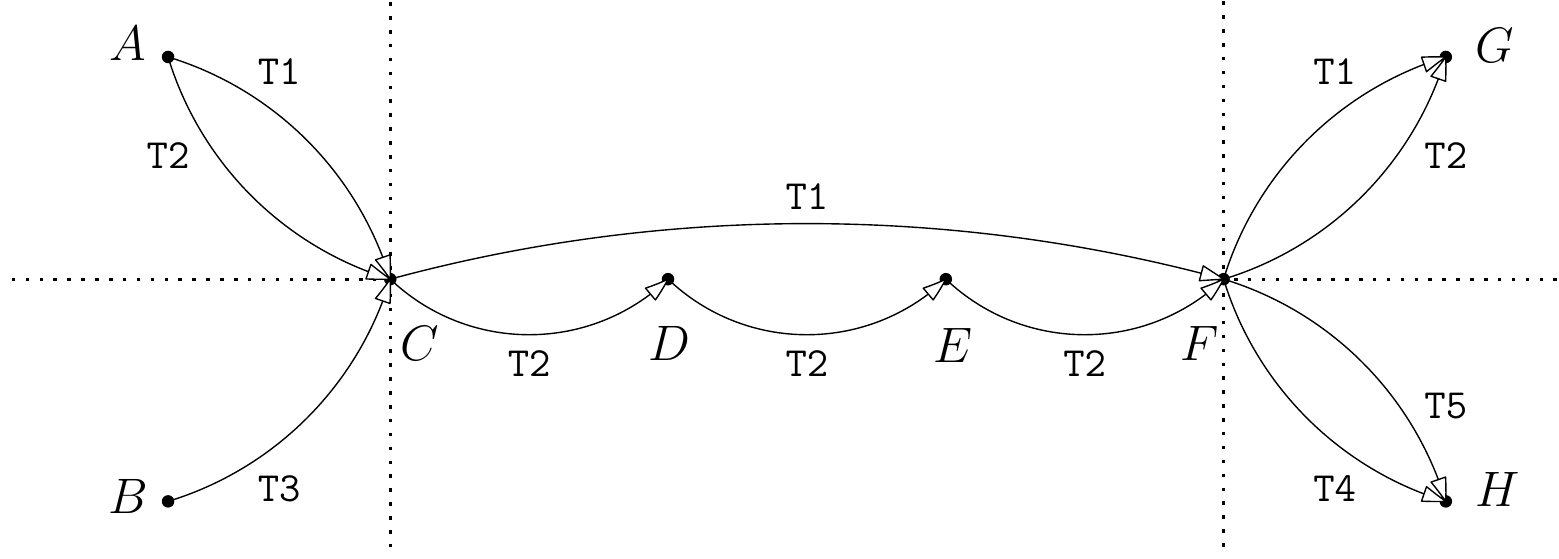}
\caption{The full domain with services from the relevant timetable. 
There are five different trains \textbf{T1} to \textbf{T5},
and train \textbf{T1} is a~direct train.}
\label{fig:groupPlanTimetable}
\end{figure}

\begin{table*}
\centering
\caption{Parameters of the testing scenarios}
\label{tab:tblScenarios}
\begin{tabular}{|l|r|r|r|r|r|} \hline
{\bf Scenario code}&{\bf S1}&{\bf S2}&{\bf S3}&{\bf S4}&{\bf S5}\\ \hline
Number of stops &
344 & 721 & $1\:044$ &  $1\:670$ &   $2\:176$\\ \hline
Relaxed domain connections &
744 & $1\:520$ &  $2\:275$ & $4\:001$ & $4\:794$\\ \hline
Timetabled connections & \quad$23\:994$ & $26\:702$ & \quad$68\:597$ &
$72\:937$ & \quad$203\:590$\\ \hline
Means of transport &
trains & trains, coaches & trains & trains, coaches & trains\\
\hline\end{tabular}
\end{table*}

In order to use both direct and stopping trains when the group journey is
matched to the timetable, the {\it relevant timetable} for a~group journey is
composed in the following way: for every part of the group journey, return all timetable
services in the direction of agents' journeys which connect the stops in that
part. An~example of the relevant timetable for a~group of agents from the
previous example is shown in Figure~\ref{fig:groupPlanTimetable}. Now,
the agents can travel using the direct train \verb|T1| or using train
\verb|T2| with intermediate stops.

The relevant timetable for the group journey is used with the aim to cut down
the amount of data that will be given to a~temporal single-agent planner. For
instance, there are $23\,994$ timetabled connections in Scotland. For an~example
journey of two agents, there are only 885 services in the relevant timetable
which is approximately 4~\% of the data. As a~result, the temporal single-agent
planner gets only the necessary amount of data as input, to prevent the
time-consuming exploration of irrelevant regions of the state space.





\subsection{Cost functions} \label{sec:costFunctions}

The timetable data used in this paper (cf.~section~\ref{sec:sourceData})
contains neither information about ticket prices nor distances between
adjacent stops, only durations of journeys from one stop
to another. This significantly restricts the design of cost functions used for
the planning problems. Therefore, the cost functions used in the three phases of
the algorithm are based solely on the duration of journeys.

In the initial phase, every agent tries to get to its destination in the
shortest possible time. The cost of travelling between adjacent stops $A$ and
$B$ is simply the duration of the journey between stops $A$ and $B$.
In the BR phase, we design the cost function in such a~way that it favours
shared journeys. 
The cost $c_{i,n}$ for agent~$i$ travelling from $A$ to $B$ in a~group of $n$ agents is then defined by equation~\eqref{eq:defGroupCost}: 
\begin{equation} \label{eq:defGroupCost} c_{i, n} = \left( \frac{1}{n}\,0.8 + 0.2 \right) c_i \end{equation}

\noindent where $c_i$ is the individual cost of the single action
to $i$ when travelling alone. In our experiments below, we take this to be equal
to the duration of the trip from $A$ to $B$.



This is designed to approximately model  the discount for the passengers if they
buy a~group ticket: The more agents travel together, the cheaper the shared (leg
of a) journey becomes for each agent.
Also, an~agent cannot travel any cheaper than 20 \% of the single-agent cost. 
In reality, pricing for group tickets could vary, and while our experimental
results assume this specific setup, the actual price calculation could be easily
replaced by any alternative model.


In the timetabling phase, every agent in a group of agents tries to spend the
shortest possible time on its journey. When matching the plan to the timetable,
the temporal planner tries to minimise the sum of durations of agents' journeys
including waiting times between services.

\section{Evaluation} \label{sec:evaluation}
We have evaluated our algorithm on public transportation data for the United
Kingdom, using various off-the-shelf planners for the three phases described
above, and a number of different scenarios. These are described together with
the results obtained from extensive experiments below.


\subsection{Planners}


All three single-agent planners used for the evaluation were taken from recent
International Planning Competitions (IPC) from 2008 and 2011. 
We use LAMA \cite{richter2008} in the initial and the BR phase, a~sequential
{\em satisficing} (as opposed to cost-optimal) planner which searches for any
plan that solves a~given problem and does  not guarantee optimality of the plans
computed. LAMA is a~propositional planning system based on heuristic state-space
search. Its core feature is the usage of landmarks
\cite{richter2008b}, i.e., propositions that must be true in every solution of
a~planning problem.

\sgp{} \cite{hsu2008} and POPF2 \cite{coles2011a} are temporal satisficing
planners used in the timetabling phase.
Such temporal planners take the duration of actions into account and try to
minimise makespan (i.e., total duration) of a~plan but do not guarantee
optimality. \jh{The two planners use different search strategies and usually
produce different results.} This allows us to run them in sequence on every
problem and to pick the plan with the shortest duration. It is not strictly necessary to
run both planners, one could save computation effort by trusting one of them.

\sgp{} consists of three inter-related steps: parallel decomposition, constraint
resolution and subproblem solution \cite{chen2006,hoffmann2001,meuleau1998, wah2006}. POPF2 is
a temporal forward-chaining partial-order planner with a specific extended
grounded search strategy described in \cite{coles2011, coles2010}.
It is not known beforehand which of the two planners will return a~better plan
on a particular problem instance.




\subsection{Scenarios}
To test the performance of our algorithm, we generated five different scenarios
of increasing complexity, whose parameters are shown in
Table~\ref{tab:tblScenarios}.
They are based on different regions of the United Kingdom (Scotland for S1 and
S2, central UK for S3 and S4, central and southern UK for S5). Each scenario
assumes trains or trains and coaches as available means of transportation.


In order to observe the behaviour of the algorithm with different numbers of
agents, we ran our algorithm on every scenario with $2, 4, 6, \dots, 14$ agents
in it. To ensure a~reasonable likelihood of travel sharing to occur, all agents
in the scenarios travel in the same direction. This imitates a~preprocessing
step where the agents' origins and destinations are clustered according to their
direction of travel. \jh{For simplicity reasons, we have chosen 
directions based on cardinal points (N--S, \hbox{S--N}, W--E, E--W).}
For every scenario and number of agents, we generated 40~different experiments
(10~experiments for each direction of travel), resulting in a~total
of  $1,400$ experiments. All experiments are generated partially randomly as
defined below.



To explain how each experiment is set up, assume we have selected a scenario
from S1 to S5, a specific number of agents,
 and a direction of travel, say north--south. To compute
the origin--destination pairs to be used by the agents, we place two axes $x$
and $y$ over the region, dividing the stops in the scenario into four quadrants
{\bf I}, {\bf II}, {\bf III} and {\bf IV}.
Then, the set~$O$ of possible
origin--destination pairs is computed as 
\begin{multline*} \label{eq:defODset}
O := \{ ( A, B ) \vert (\left (A \in \mathbf{I} \wedge B \in \mathbf{IV}
\right ) \vee  \left ( A \in \mathbf{II} \wedge B
\in \mathbf{III} \right )) \\ \wedge \vert AB \vert \in [ 20, 160] \}
\end{multline*}
This means that each agent travels from $A$ to $B$ either from 
quadrant {\bf I} to {\bf IV} or from quadrant {\bf II} to {\bf III}. The
straight-line distance $\left\vert AB \right\vert$ between the origin and the
destination is taken from the interval 20--160~km (when using roads or
rail tracks, this interval stretches approximately to a real distance of
30--250~km). This interval is chosen to prevent journeys that could be hard
to complete within 24~hours. We sample the actual origin-destination pairs 
from the elements of $O$, assuming a uniform distribution, and repeat the 
process for all other directions of travel.

\subsection{Experimental results}
We evaluate the performance of the algorithm in terms of three different
metrics: the amount of time the algorithm needs to compute shared journeys for
all agents in a~given scenario, the success rate of finding a~plan for any given
agent and the quality of the plans computed. Unless stated otherwise, the values
in graphs are averaged over 40~experiments that were performed for each
scenario and each number of agents. The results obtained are based on running
the algorithm on a Linux desktop computer with 2.66~GHz Intel Core~2 Duo
processor and 4~GB of memory. \jhhh{The data, source codes and scenarios in PDDL
are archived {online}\footnote{
\href{http://agents.fel.cvut.cz/download/hrncir/journey_sharing.tgz}{agents.fel.cvut.cz/download/hrncir/journey\_sharing.tgz}}.}


\subsubsection{Scalability}
To assess the scalability of the algorithm, we measure the amount of time
needed to plan shared journeys for all agents in a~scenario.

In many of the experiments, the \sgp{} and POPF2 used in the timetabling phase
returned some plans in the first few minutes but then they continued exploration
of the search space without returning any better plan.
To account for this, \jh{we imposed a~time limit for each planner in the
temporal planning stage} to 5 minutes for a group of up to 5 agents, 10 minutes
for a group of up to 10 agents, and 15 minutes otherwise.

\begin{figure}[t!]
\centering
\includegraphics[width=74mm]{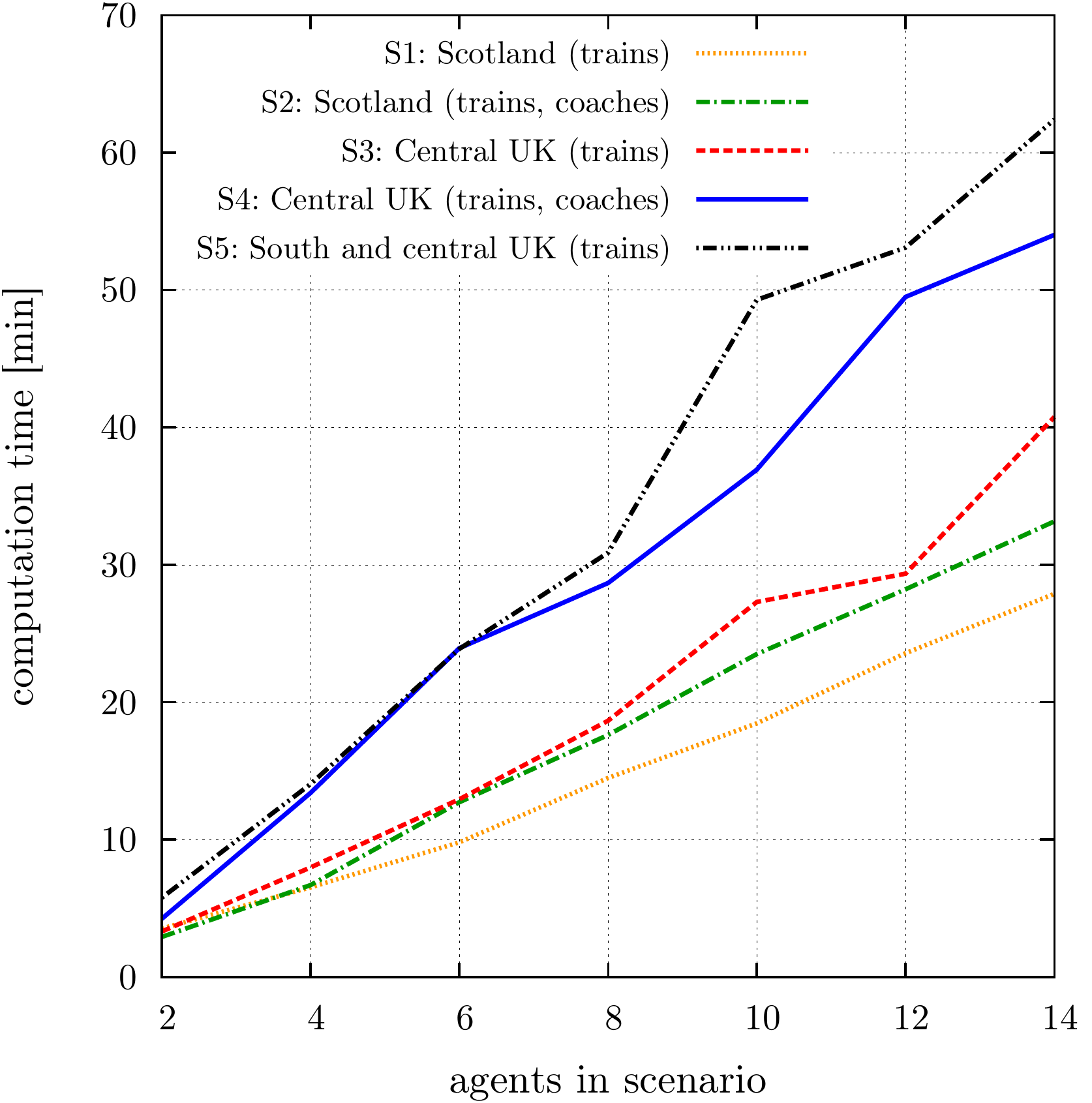}
\caption{Computation time against number of agents}
\label{fig:runtime}
\end{figure}

\begin{figure}[t!]
\centering
\includegraphics[width=71mm]{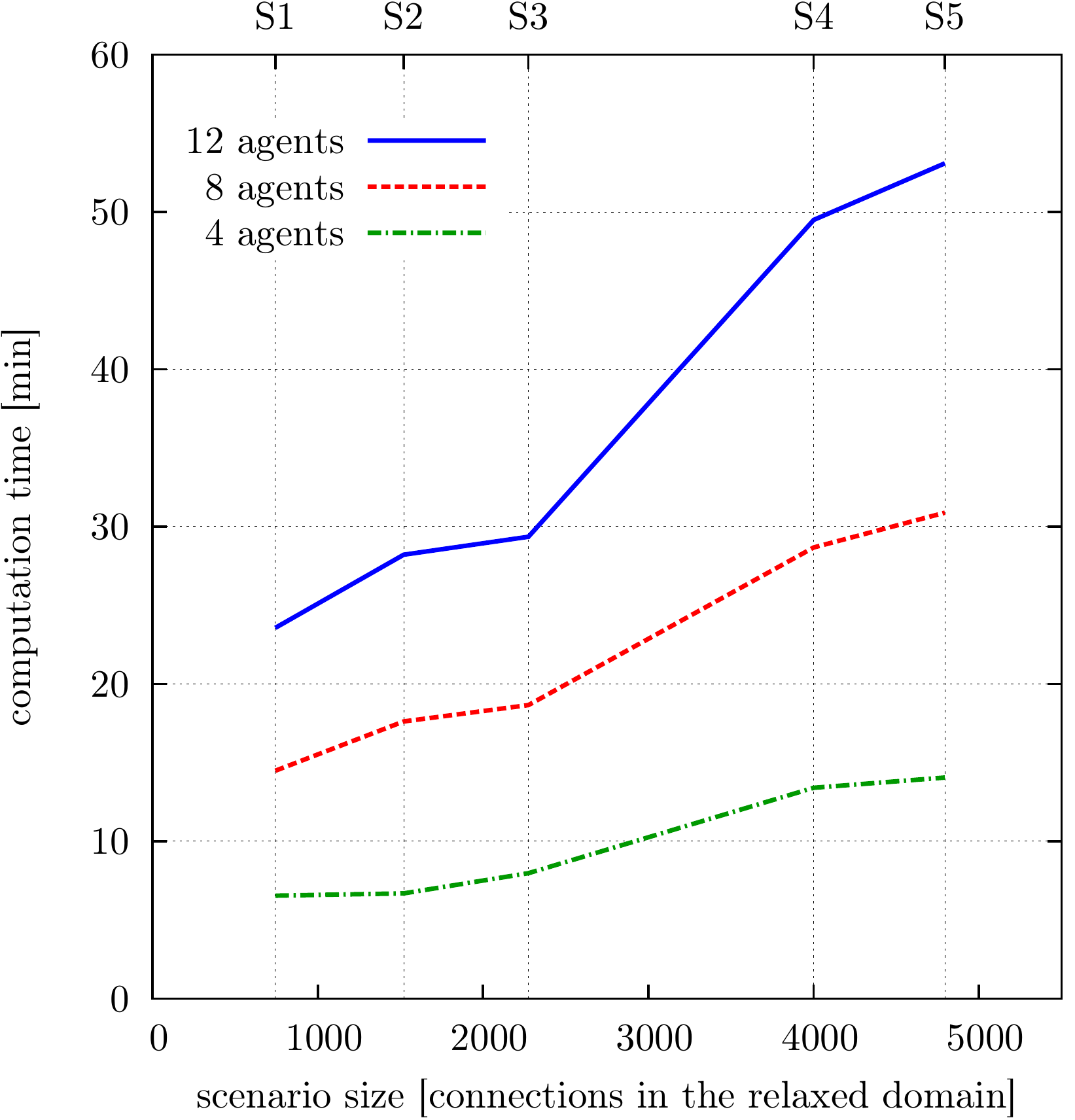}
\caption{Computation time against scenario size}
\label{fig:runtimeSize2}
\end{figure}

Figure~\ref{fig:runtime} shows the computation times of the algorithm.
The graph indicates that overall computation time grows roughly linearly with
increasing number of agents, which confirms that the algorithm avoids the
exponential blowup in the action space characteristic for centralised multiagent
planning.
Computation time also increases linearly with growing scenario size.
Figure~\ref{fig:runtimeSize2} shows computation times for 4, 8 and 12 agents
against the different scenarios.

While the overall computation times are considerable (up to one hour for 14
agents in the largest scenario), we should emphasise that the algorithm is
effectively computing equilibrium solutions in multi-player games with hundreds
of thousands of states. Considering this, the linear growth hints at having
achieved a level of scalability based on the structure of the domain that is far
above naive approaches to plan jointly in such state spaces. Moreover, it
implies that the runtimes could be easily reduced by using more processing
power.

\subsubsection{Success rate}
To assess the value of the algorithm, we also need to look at how many agents
end up having a valid travel plan.
\jh{Planning in the relaxed domain in the initial and the BR phase of the
algorithm is very successful. After the BR phase, 99.4~\% of agents have a~journey plan.
The remaining 0.6~\% of all agents does not have a single-agent plan because of
the irregularities in the relaxed domain caused by splitting the public
transportation network into regions.
The agents without a single-agent plan are not matched to timetable connections
in the timetabling phase.}


\begin{figure}[t!]
\centering
\includegraphics[width=76mm]{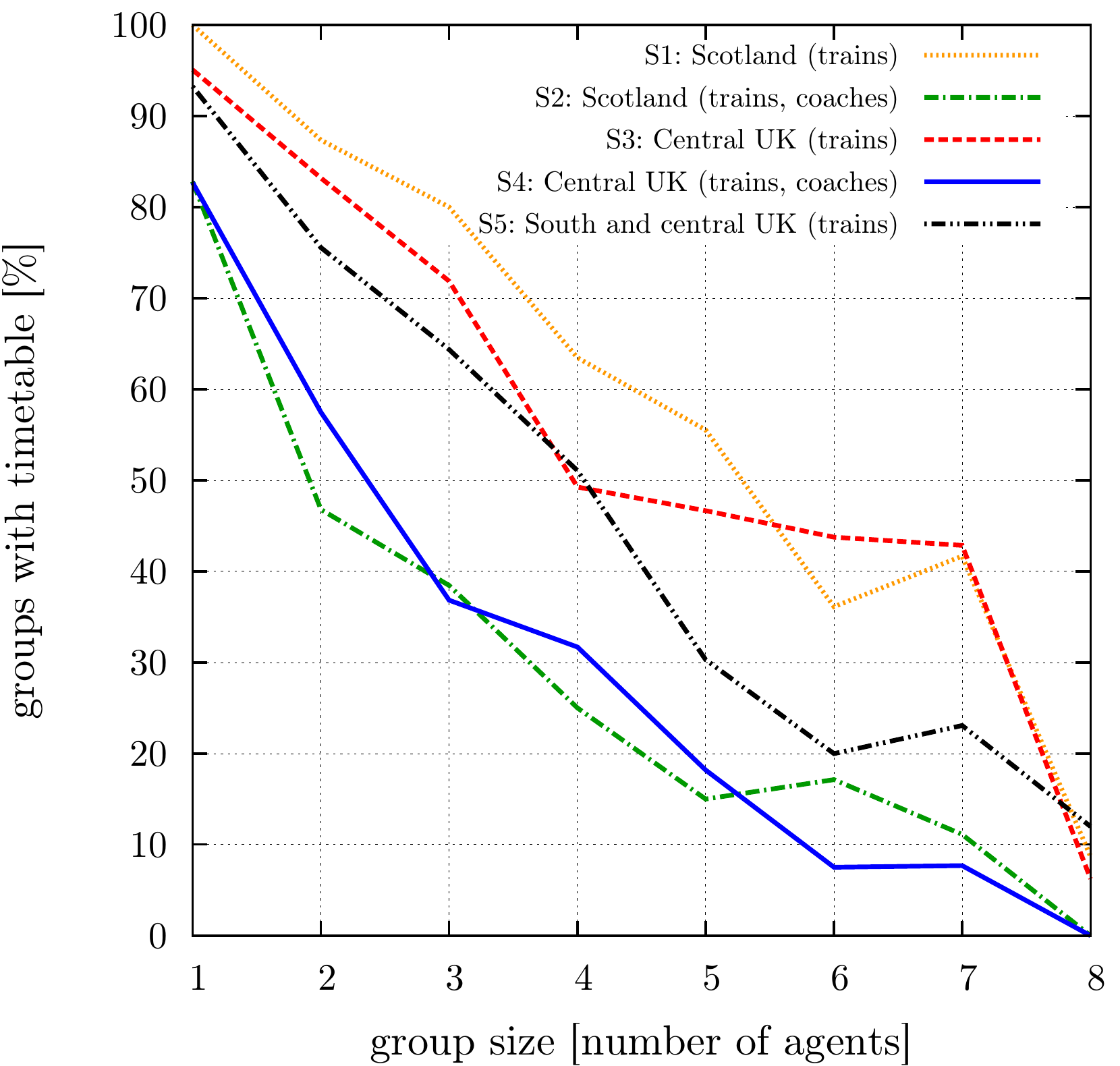}
\caption{Percentage of groups for which a~timetable was
found as a function of group size.}
\label{fig:groupsWithTimetable}
\end{figure}

The timetabling phase is of course much more problematic.
Figure~\ref{fig:groupsWithTimetable} shows the percentage of groups for which
a~timetable was found, as a function of group size.
\jh{In order to create this graph, number of groups with assigned timetable and
total number of groups identified was counted for every size of the group.}
There are several things to point out here.

Naturally, the bigger a group is, the harder it is to find a~feasible timetable,
as the problem quickly becomes overconstrained in terms of travel times and
actually available transportation services. When a~group of agents sharing parts
of their journeys is big (5~or more agents), the percentage of groups for which
we can find a timetable drops below 50~\%. With a~group of 8~agents, almost no
timetable can be found. Basically what happens here is that the initial and BR
phases find suitable ways of travelling together in principle, but that it
becomes impossible to find appropriate connections that satisfy every
traveller's requirements, \jh{or do not add up to a total duration of less than
24~hours.}


We can also observe that the success rate is higher in scenarios that use only
trains than in those that combine trains and coaches.
On closer inspection, we can observe that this is mainly caused by different
{\em service densities} in the rail and coach networks, i.e., the ratios of
timetabled connections over connections in the relaxed domain. For example, the
service density is 33~train services a~day compared to only 4~coach services  in
Scotland. As a~consequence, it is much harder to find a~timetable in a~scenario
with both trains and coaches because the timetable of coaches is much less
regular than the timetable of trains. However, this does not mean that there is
less sharing if coaches are included. Instead, it just reflects the fact that
due to low service density, many of the envisioned shared journeys do not turn
out to be feasible using coaches. The fact that this cannot be anticipated in
the initial and BR phases is a weakness of our method, and is discussed further
in section~\ref{sec:conclusion}.

\begin{figure}[t!]
\centering
\includegraphics[width=76mm]{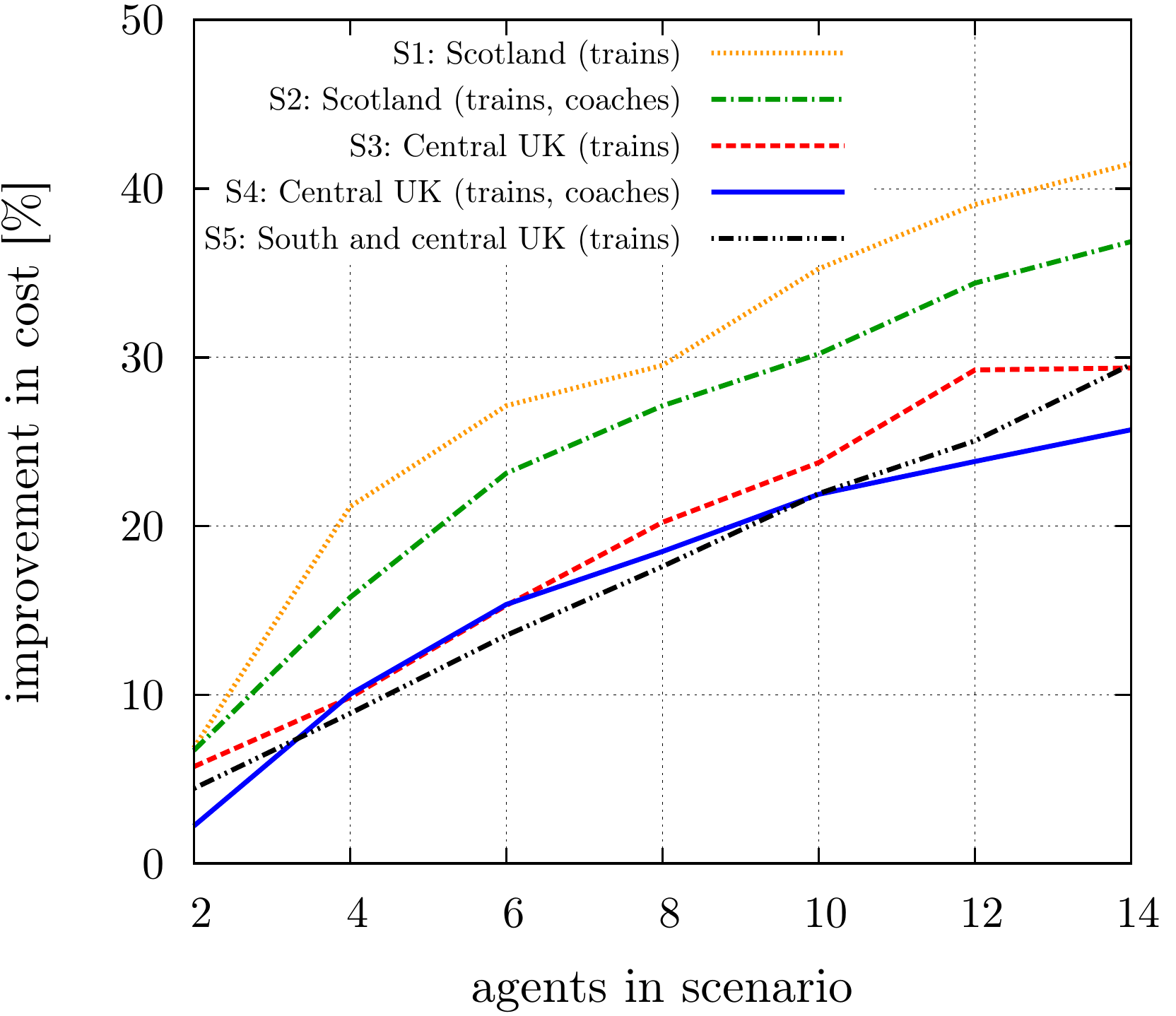}
\caption{Average cost improvement}
\label{fig:improvement}
\end{figure}

\begin{figure}[t!]
\centering
\includegraphics[width=73mm]{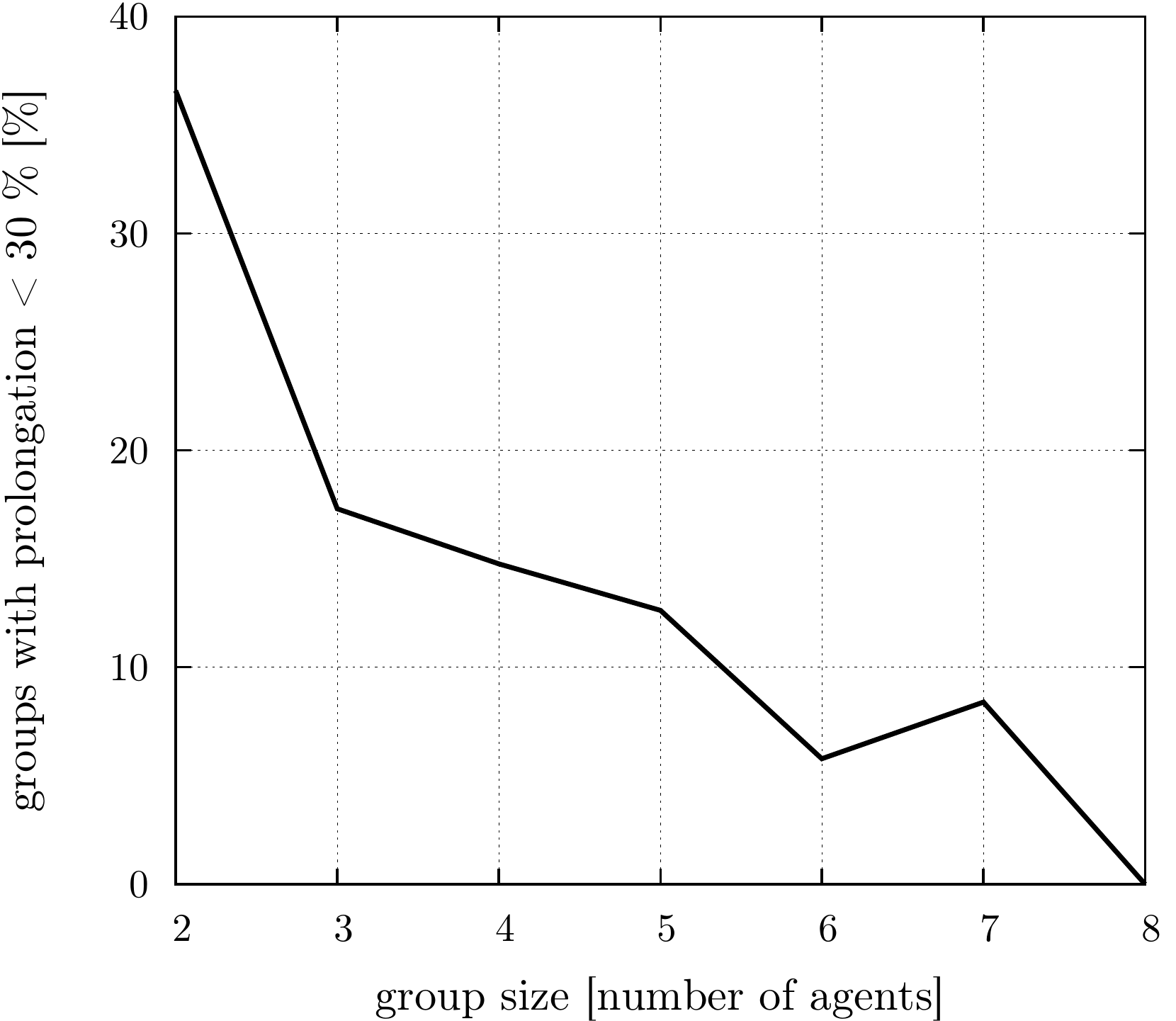}
\caption{Percentage of groups with less than 30~\% journey prolongation}
\label{fig:prolongationLimitGlobal}
\end{figure}

\subsubsection{Plan quality}

Finally, we want to assess the quality of the plans obtained with respect to
improvement in cost of agents' journeys and their prolongation, to evaluate the
net benefit of using our method in the travel sharing domain. \jhh{We should
mention that the algorithm does not explicitly optimises the solutions with
respect to these metrics.}
To calculate cost improvement, recalling that $C_{i}(\pi) = \sum_j c_i(a^j)$ for
a plan is the cost of a plan $\pi=\langle a^1,\ldots, a^k\rangle$ to agent $i$,
assume $n(a^j)$ returns the number of agents with whom the $j$th step of the
plan is shared.
We can define \jh{a~cost of a~shared travel plan} $C_{i}^{'}(\pi)=\sum_j
c_{i,n(a^j)}(a^j)$ using equation~\eqref{eq:defGroupCost}.
\jhc{I added $'$ to $C_{i}$ to distinguish it from the cost of non-shared
travel.} \jhd{, where $c_{i,n(a^j)}(a^j)=c_i(a^j)$ if
$n(a^j)=0$. }\jhc{deleted because if there is only one agent in a group,
the formula (1) just returns the single-agent cost}
With this we can calculate the improvement $\Delta C$ as follows:
\begin{equation} \label{eq:defImprovement} \Delta C = \frac{\sum_{i\in N}
C_{i}(\pi_i) - \sum_{i\in N} C_{i}^{'}(\pi_N)}{\sum_{i\in N} C_{i}(\pi_i)}
\end{equation}
where $N$ is the set of all agents, $\pi_i$ is the single-agent plan
initially computed for agent $i$, and $\pi_N$ is the final joint plan of all
agents after completion of the algorithm (which is interpreted as the plan
of the ``grand coalition'' $N$ and reflects how subgroups within $N$ share parts
of the individual journeys).

%

The average cost improvement obtained in our experiments is shown in
Figure~\ref{fig:improvement}, and it shows that the more agents there are in the
scenario, the higher the improvement. However, there is a~trade-off between the
improvement in cost and the percentage of groups that we manage to find a
suitable timetable for, cf.~Figure~\ref{fig:groupsWithTimetable}.

On the one hand, travel sharing is beneficial in terms of cost. On the other
hand, a~shared journey has a longer duration than a~single-agent journey in most
cases. In order to evaluate this trade-off, we also measure the journey
prolongation.
Assume that $T_{\mathrm{i}}(\pi)$ is the total duration of a plan to agent $i$
in plan $\pi$, and, as above, $\pi_i$/$\pi_N$ denote the initial single-agent
plans and the shared joint plan at the end of the timetabling phase,
respectively. Then, the prolongation $\Delta T$ of a~journey is defined as
follows: 
\begin{equation} \label{eq:defProlongation} \Delta T = \frac{\sum_{i\in N}
T_i(\pi_N) - \sum_{i\in N} T_i(\pi_i)}{\sum_{i\in N} T_i(\pi_i)}
\end{equation}
Journey prolongation can be calculated only when a~group is assigned a~timetable
and each member of the group is assigned a~single-agent timetable. For this
purpose, in every experiment, we also calculate single-agent timetables in the
timetabling phase of the algorithm.

A~graph of the percentage of groups that have a~timetable with prolongation less
than 30 \% as a function of group size is shown in
Figure~\ref{fig:prolongationLimitGlobal}. The graph shows which groups benefit
from travel sharing, i.e., groups whose journeys are not prolonged excessively
by travelling together.
Approximately 15~\% of groups with 3--4 agents
are assigned a~timetable that leads to a prolongation of less than~30~\%. Such a~low
percentage of groups can be explained by the algorithm trying to optimise the
price of the journey by sharing in the BR phase. However, there is a~trade-off
between the price and the duration of the journey. The more agents are sharing
a~journey, the longer the journey duration is likely to be.

\jh{These results were obtained based on the specific cost
function~\eqref{eq:defGroupCost} we have introduced to favour travel sharing},
and which would have to be adapted to the specific cost structure that is
present in a given transportation system. Also, the extent to which longer
journey times are acceptable for the traveller depends on their preferences, but
these could be easily adapted by using different cost functions.





\section{Discussion} \label{sec:discussion}
The computation of single-agent plans in the initial phase involves solving a
set of completely independent planning problems. This means that the planning
process could be speeded up significantly by using parallel computation on
multiple CPUs. The same is true for matching different independent groups of
agents to timetabled connections in the timetabling phase. As an~example, assume that
there are $N$~agents in the scenario and $t_1, \dots, t_N$ are the computation
times for respective single-agent initial plans. If computed concurrently, this
would reduce the computation time from $t = \sum_{i=1}^N t_i$ to $t' =
\max_{i=1}^N (t_i)$. Similar optimisations could be performed for the
timetabling phase of the algorithm.
In the experiments with 10~agents, for example, this would lead to a~runtime
reduced by 48~\% in scenario~S1 and by 44~\% in scenario~S5.


A major problem of our method is the inability to find appropriate connections
in the timetabling phase for larger groups.
There are several reasons for this. Firstly, the relaxed domain is overly
simplified, and many journeys found in it do not correspond to journeys that
would be found if we were planning in the full domain. Secondly, there are too
many temporal constraints in bigger groups (5 or more agents), so the timetable
matching problem becomes unsolvable given the 24-hour timetable.
\jh{However, it should also be pointed out that such larger groups would be
very hard to identify and schedule even using human planning.}
Thirdly, some parts of public transportation network have very irregular
timetables.

Our method clearly improves the cost of agents' journeys by sharing parts of the
journeys, even though there is a~trade-off between the amount of improvement,
the percentage of found timetables and the prolongation of journeys. On the one
hand, the bigger the group, the better the improvement. On the other hand, the
more agents share a~journey, the harder it is to match their joint plan to
timetable. Also, the prolongation is likely to be higher with more agents
travelling together, and will most likely lead to results that are not
acceptable for users in larger groups.


Regarding the domain-independence of the algorithm, we should point out that its
initial and BR~phases are completely domain-independent so they could easily be
used in other problem domains such as logistics, network routing or service
allocation. In the traffic domain, the algorithm can be used to plan routes that
avoid traffic jams or to control traffic lights. What is more, additional
constraints such as staying at one city for some time or travelling together
with a~specific person can be easily added. On the other hand, the timetabling
phase of the algorithm is domain-specific, providing an example of the specific
design choices that have to be made from an engineering point of view.



To assess the practical value of our contribution, it is worth discussing how it
could be used in practice as a~part of a~travel planning system for real
passengers. In such a~system, every user would submit origin, destination and
travel times.
Different users could submit their preferences at different times, with  the
system continuously computing shared journeys for them based on information
about all users' preferences.
Users would need to agree on a shared journey in time to arrange meeting points
and to purchase tickets, subject to any restrictions on advance tickets etc.
Because of this lead time, it would be entirely sufficient if the users got an
e-mail with a~planned journey one hour after the last member of the travel group
submits his or her journey details, which implies that even with our current
implementation of the algorithm, the runtimes would be acceptable.


From our experimental evaluation, we conclude that reasonable group sizes range
from two to four persons. Apart from the fact that such groups can be relatively
easily coordinated, with the price model used in this paper,
cf.~formula~\eqref{eq:defGroupCost}, every member of a~three-person group could
save up to 53~\% of the single-agent price. The success rate of the timetabling
phase of the algorithm for three-person groups in the scenario S3 (trains in the
central UK) is 70 \%.

\section{Conclusion} \label{sec:conclusion}


We have presented a~multiagent planning algorithm which is able to plan
meaningful shared routes in a~real-world travel domain. The algorithm has been
implemented and evaluated on five scenarios based on real-world UK~public
transport data. The algorithm exhibits very good scalability, since it scales linearly
both with the scenario size and the number of agents. The average computation
time for 12~agents in the scenario with 90~\% of trains in the UK is less than
one hour. Experiments indicate that the algorithm avoids the exponential blowup
in the action space characteristic for a~centralised multiagent planner.

To deal with thousands of users that could be in a~real-world travel planning
system, a~preprocessing step would be needed: The agents would have to be
divided into smaller groups by clustering them according to departure time,
direction of travel, origin, destination, length of journey and preferences
(e.g., travel by train only, find cheapest journey). Then, the algorithm could
be used to find a~shared travel plan with a~timetable. To prevent too large
groups of agents which are unlikely to be matched to the timetable, a~limit can
be imposed on the size of the group.
If a~group plan cannot be mapped to a timetable, the group can be split into
smaller sub-groups which are more likely to identify a~suitable timetable.



Finally, the price of travel and flexibility of travel sharing can be
significantly improved by sharing a~private car. In the future, we would like to
explore the problem of planning shared journeys when public transport is
combined with ride sharing. Then, in order to have a~feasible number of nodes in
the travel domain, train and bus stops can be used as meeting points where it is
possible to change from a~car to public transport or vice versa.

\section{Acknowledgments}



Partly supported by the Ministry of Education, Youth and Sports of Czech
Republic (grant No. LD12044) and European Commission FP7 (grant agreement No.
289067).

%
\bibliographystyle{abbrv}
\bibliography{references}
%
%

%

\end{document}